\documentclass[runningheads]{llncs}
\usepackage{graphicx}
\usepackage{xcolor}
\usepackage{hyperref}
\usepackage{multirow}
\usepackage{times}
\usepackage{enumitem}
\usepackage{subcaption}

\DeclareCaptionSubType*[arabic]{table}
\begin{document}
\title{FacTweet: Profiling Fake News Twitter Accounts}
%

\author{Bilal Ghanem\inst{1} \and
Simone Paolo Ponzetto\inst{2} \and
Paolo Rosso\inst{1}}
\authorrunning{B. Ghanem et al.}
\institute{
PRHLT Research Center, Universitat Polit\`ecnica de Val\`encia, Spain \\
\email{\{bigha@doctor, prosso@dsic\}.upv.es} \and
University of Mannheim, Germany \\
\email{simone@informatik.uni-mannheim.de}
}

\maketitle              

\begin{abstract}
We present an approach to detect fake news in Twitter at the account level using a neural recurrent model and a variety of different semantic and stylistic features.
%
%
%
Our method extracts a set of features from the timelines of news Twitter accounts by reading their posts as chunks, rather than dealing with each tweet independently. We show the experimental benefits of modeling latent stylistic signatures of mixed fake and real news with a sequential model over a wide range of strong baselines.
\end{abstract}

\section{Introduction}
\label{sec:introduction}
Social media platforms have made the spreading of fake news easier, faster as well as able to reach a wider audience. Social media offer another feature which is the anonymity for the authors, and this opens the door to many suspicious individuals or organizations to utilize these platforms. Recently, there has been an increased number of spreading fake news and rumors over the web and social media \cite{vosoughi2018spread}. 

Fake news in social media vary considering the intention to mislead. Some of these news are spread with the intention to be ironic or to deliver the news in an ironic way (satirical news). Others, such as propaganda, hoaxes, and clickbaits, are spread to mislead the audience or to manipulate their opinions. In the case of Twitter, suspicious news annotations should be done on a tweet rather than an account level, since some accounts mix fake with real news. However, these annotations are extremely costly and time consuming -- i.e., due to high volume of available tweets Consequently, a first step in this direction, e.g., as a pre-filtering step, can be viewed as the task of detecting fake news at the account level. 

The main obstacle for detecting suspicious Twitter accounts is due to the behavior of mixing some real news with the misleading ones. Consequently, we investigate ways to detect suspicious accounts by considering their tweets in groups (chunks). Our hypothesis is that suspicious accounts have a unique pattern in posting tweet sequences. Since their intention is to mislead, the way they transition from one set of tweets to the next has a hidden signature, biased by their intentions. Therefore, reading these tweets in chunks has the potential to improve the detection of the fake news accounts.

In this work, we investigate the problem of discriminating between factual and non-factual accounts in Twitter. To this end, we collect a large dataset of tweets using a list of \emph{propaganda}, \emph{hoax} and \emph{clickbait} accounts and compare different versions of sequential chunk-based approaches using a variety of feature sets against several baselines. Several approaches have been proposed for news verification, whether in social media (rumors detection) \cite{vosoughi2018spread,volkova2017separating,ma2016detecting,zhao2015enquiring,castillo2011information}, or in news claims \cite{karadzhov2017fully,li2011t,ghanem2018stance,baly2018predicting}. The main orientation in the previous works is to verify the textual claims/tweets but not their sources. To the best of our knowledge, this is the first work aiming to detect factuality at the account level, and especially from a textual perspective. 
Our contributions are:

\begin{itemize}[leftmargin=4mm]
   \item We propose an approach to detect non-factual Twitter accounts by treating post streams as a sequence of tweets' chunks. We test several semantic and dictionary-based features together with a neural sequential approach, and apply an ablation test to investigate their contribution.
   
   \item We benchmark our approach against other approaches that discard the chronological order of the tweets or read the tweets individually. The results show that our approach produces superior results at detecting non-factual accounts.
\end{itemize}

\section{Methodology}
Given a news Twitter account, we read its tweets from the account's timeline. Then we sort the tweets by the posting date in ascending way and we split them into $N$ chunks. 
Each chunk consists of a sorted sequence of tweets labeled by the label of its corresponding account. We extract a set of features from each chunk and we feed them into a recurrent neural network to model the sequential flow of the chunks' tweets. We use an attention layer with dropout to attend over the most important tweets in each chunk. Finally, the representation is fed into a softmax layer to produce a probability distribution over the account types and thus predict the factuality of the accounts. Since we have many chunks for each account, the label for an account is obtained by taking the majority class of the account's chunks.

\vspace{1em}
\noindent \textbf{Input Representation.} Let $t$ be a Twitter account that contains $m$ tweets. These tweets are sorted by date and split into a sequence of chunks $ck = \langle ck_1, \ldots, ck_n \rangle$, where each $ck_i$ contains $s$ tweets. Each tweet in $ck_i$ is represented by a vector $v \in {\rm I\!R}^d$ , where $v$ is the concatenation of a set of features' vectors, that is $v = \langle f_1, \ldots, f_n \rangle$. Each feature vector $f_i$ is built by counting the presence of tweet's words in a set of lexical lists. The final representation of the tweet is built by averaging the single word vectors.

\vspace{1em}
\noindent \textbf{Features.} We argue that different kinds of features like the sentiment of the text, morality, and other text-based features are critical to detect the nonfactual Twitter accounts by utilizing their occurrence during reporting the news in an account's timeline. We employ a rich set of features borrowed from previous works in fake news, bias, and rumors detection \cite{vosoughi2018spread,volkova2017separating,baly2018predicting,horne2017just}.

\begin{itemize}[leftmargin=4mm]
    \item \textbf{Emotion}: We build an emotions vector using word occurrences of 15 emotion types from two available emotional lexicons. We use the NRC lexicon \cite{mohammad2010emotions}, which contains $\sim$14K words labeled using the eight Plutchik's emotions \cite{plutchik1980general}. The other lexicon is SentiSense \cite{de2012sentisense} which is a concept-based affective lexicon that attaches emotional meanings to concepts from the WordNet\footnote{\url{https://wordnet.princeton.edu}} lexical database. It has $\sim$5.5 words labeled with emotions from a set of 14 emotional categories 
    We use the categories that do not exist in the NRC lexicon. 
    
    \item \textbf{Sentiment}: We extract the sentiment of the tweets by employing EffectWordNet \cite{choi2014+}, SenticNet \cite{cambria2014senticnet}, NRC \cite{mohammad2010emotions}\footnote{NRC has also two sentiment categories, positive and negative.}, and subj\_lexicon \cite{wilson2005recognizing}, where each has the two sentiment classes, \textit{positive} and \textit{negative}. 
    
    
    \item \textbf{Morality}: Features based on morality foundation theory \cite{graham2009liberals} where words are labeled in one of the following 10 categories (\textit{care, harm, fairness, cheating, loyalty, betrayal, authority, subversion, sanctity,} and \textit{degradation}). 
    
    \item \textbf{Style}: We use canonical stylistic features, such as the count of question marks, exclamation marks, consecutive characters and letters\footnote{We considered 2 or more consecutive characters, and 3 or more consecutive letters.}, links, hashtags, users' mentions. In addition, we extract the uppercase ratio and the tweet length. 
    
    \item \textbf{Words embeddings}: We extract words embeddings of the words of the tweet using $Glove\-840B-300d$ \cite{pennington2014glove} pretrained model. The tweet final representation is obtained by averaging its words embeddings. 
\end{itemize}

\begin{figure}[!tb] 
  \centering
  \includegraphics[width=5cm]{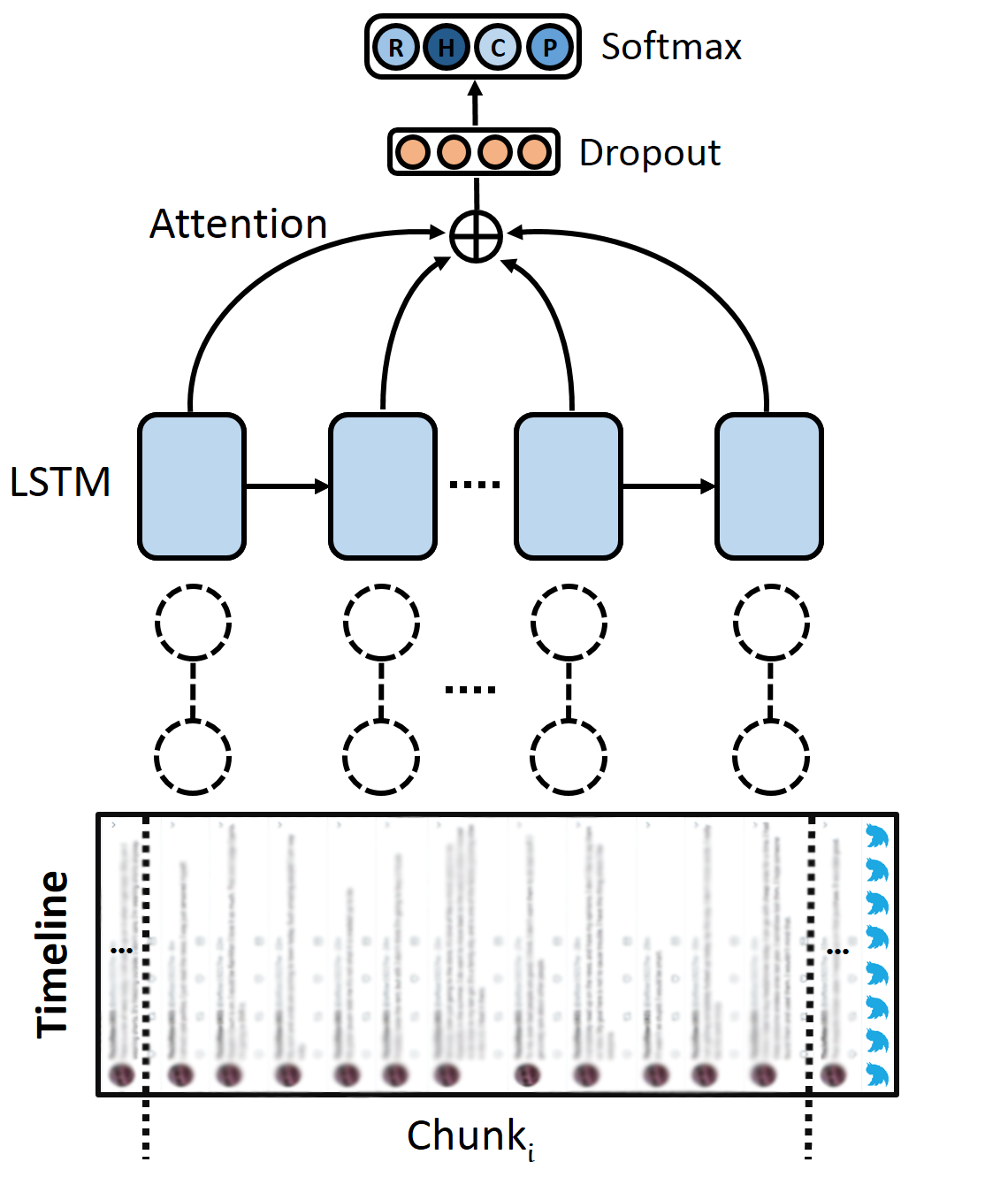}
  \caption{The FacTweet's architecture.}
  \label{fig:model}
\end{figure}

\noindent \textbf{Model}. To account for chunk sequences we make use of a \emph{de facto} standard approach and opt for a recurrent neural model using long short-term memory (LSTM) \cite{hochreiter1997long}. In our model, the sequence consists of a sequence of tweets belonging to one chunk (Figure \ref{fig:model}). The LSTM learns the hidden state $h\textsubscript{t}$ by capturing the previous timesteps (past tweets). The produced hidden state $h\textsubscript{t}$ at each time step is passed to the attention layer which computes a `context' vector $c\textsubscript{t}$ as the weighted mean of the state sequence $h$ by:
\begin{equation}
c\textsubscript{t} = \sum_{j=1}^{T} \alpha \textsubscript{tj}h\textsubscript{j},
\end{equation}
Where $T$ is the total number of timesteps in the input sequence and $\alpha \textsubscript{tj}$ is a weight computed at each time step $j$ for each state h\textsubscript{j}.

\section{Experiments and Results}

\begin{table*}[t!]
\centering
\caption{Statistics on the data with respect to each account type: propaganda (\textbf{P}), clickbait (\textbf{C}), hoax (\textbf{H}), and real news (\textbf{R}).}
\begin{tabular}{p{4cm}p{1cm}p{1cm}p{1cm}p{1cm}}
  \multirow{2}{*}{} & \multicolumn{4}{c}{Accounts Types}\\
   & P & C & H & R \\
  \hline
  \# of accounts & 96 & 36 & 7 & 32  \\
  Max \# of tweets/account & 3,250 & 3,246 & 3,250 & 3,250 \\
  Min \# of tweets/account & 33 & 877 & 453 & 212  \\
  Avg \# of tweets/account & 2,978 & 3,112 & 2,723 & 3,124 \\
  Total \# of tweets & 291,885 & 112,050 & 19,065 & 99,967 \\
\end{tabular}
\label{table:stat}
\end{table*}

\noindent \textbf{Data.} We build a dataset of Twitter accounts based on two lists annotated in previous works. For the non-factual accounts, we rely on a list of ~180 Twitter accounts from \cite{volkova2017separating}\footnote{Many of the accounts were deactivated during the collecting process, consequently only 144 accounts were used.}. This list was created based on public resources\footnote{\url{http://www.propornot.com/p/the-list.html}} where suspicious Twitter accounts were annotated with the main fake news types (clickbait, propaganda, satire, and hoax). We discard the satire labeled accounts since their intention is not to mislead or deceive. On the other hand, for the factual accounts, we use a list with another 32 Twitter accounts from \cite{karduni2018can} that are considered trustworthy by independent third parties\footnote{\url{https://tinyurl.com/yctvve9h}}. We discard some accounts that publish news in languages other than English (e.g., Russian or Arabic). Moreover, to ensure the quality of the data, we remove the duplicate, media-based, and link-only tweets. For each account, we collect the maximum amount of tweets allowed by Twitter API. Table \ref{table:stat} presents statistics on our dataset.

\begin{figure}[!tb]
   \begin{minipage}{0.47\textwidth}
        \centering
        \includegraphics[width=\textwidth]{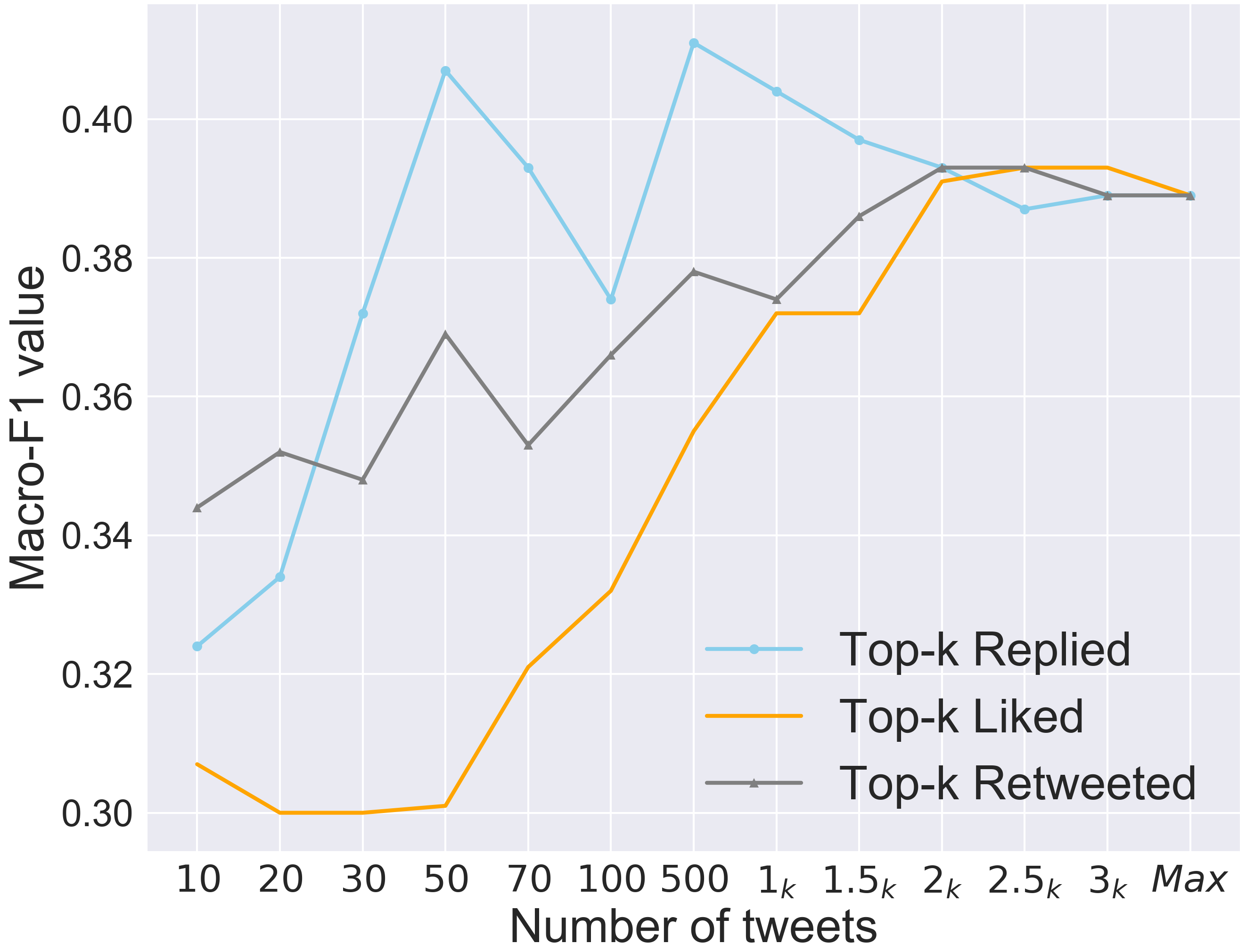}
        \caption{Results on the top-K replied, linked or re-tweeted tweets.}
        \label{fig:topk}
   \end{minipage}\vspace{5mm}\hfill
   \begin{minipage}{0.47\textwidth}
        \centering
        \includegraphics[width=\textwidth]{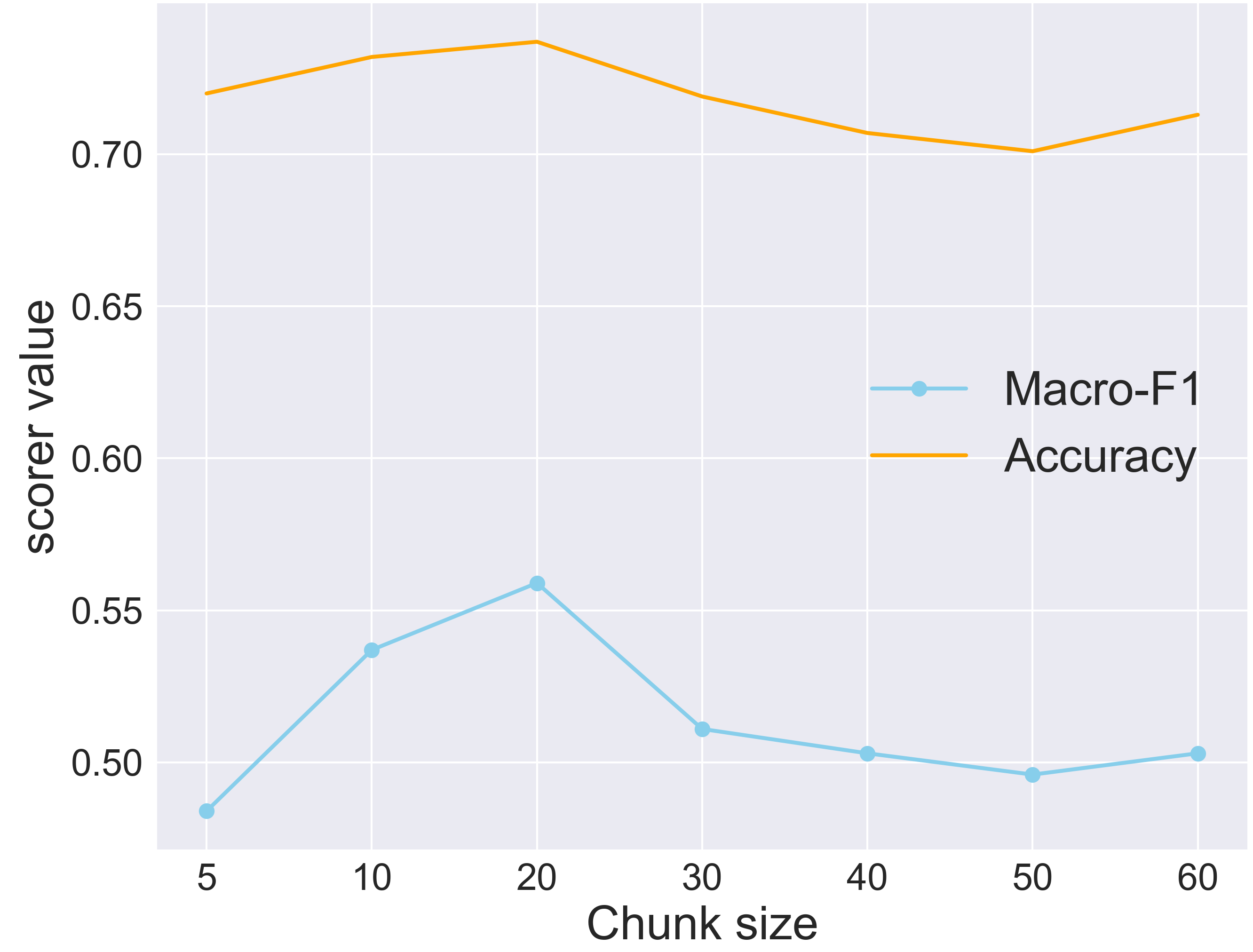}
        \caption{The FacTweet performance on difference chunk sizes.}
        \label{fig:chunk_size}
   \end{minipage}
\end{figure}

\vspace{1em}
\noindent \textbf{Baselines.} We compare our approach (FacTweet) to the following set of baselines:


\begin{itemize}[leftmargin=4mm]

\item \textbf{LR + Bag-of-words}: We aggregate the tweets of a feed and we use a bag-of-words representation with a logistic regression (LR) classifier.

\item \textbf{Tweet2vec}: We use the Bidirectional Gated recurrent neural network model proposed in \cite{dhingra2016tweet2vec}. 
We keep the default parameters that were provided with the implementation. To represent the tweets, we use the decoded embedding produced by the model. With this baseline we aim at assessing if the tweets' hashtags may help detecting the non-factual accounts.

\item \textbf{LR + All Features (tweet-level)}: We extract all our features from each tweet and feed them into a LR classifier. Here, we do not aggregate over tweets and thus view each tweet independently.

\item \textbf{LR + All Features (chunk-level)}:
We concatenate the features' vectors of the tweets in a chunk and feed them into a LR classifier.

\item \textbf{FacTweet (tweet-level)}: Similar to the FacTweet approach, but at tweet-level; the sequential flow of the tweets is not utilized. We aim at investigating the importance of the sequential flow of tweets.

\item \textbf{Top-$k$ replies, likes, or re-tweets}: Some approaches in rumors detection use the number of replies, likes, and re-tweets to detect rumors \cite{ghanem2019upv}. Thus, we extract top $k$ replied, liked or re-tweeted tweets from each account to assess the accounts factuality. We tested different $k$ values between 10 tweets to the max number of tweets from each account. Figure \ref{fig:topk} shows the macro-F1 values for different $k$ values. It seems that $k=500$ for the top \textit{replied} tweets achieves the highest result. Therefore, we consider this as a baseline.
\end{itemize}

\vspace{1em}
\noindent \textbf{Experimental Setup.} We apply a 5 cross-validation on the account's level. For the FacTweet model, we experiment with 25\% of the accounts for validation and parameters selection. We use hyperopt library\footnote{\url{https://github.com/hyperopt}} to select the hyper-parameters on the following values: LSTM layer size ($16$, $32$, $64$), dropout ($0.0-0.9$), activation function ($relu$, $selu$, $tanh$), optimizer ($sgd$, $adam$, $rmsprop$) with varying the value of the learning rate (1e{-1,..,-5}), and batch size ($4$, $8$, $16$). The validation split is extracted on the class level using stratified sampling: we took a random 25\% of the accounts from each class since the dataset is unbalanced. Discarding the classes' size in the splitting process may affect the minority classes (e.g. hoax).
For the baselines' classifier, we tested many classifiers and the LR showed the best overall performance. 

\vspace{1em}
\noindent \textbf{Results.} Table \ref{table:results} presents the results. We present the results using a chunk size of 20, which was found to be the best size on the held-out data. Figure \ref{fig:chunk_size} shows the results of different chunks sizes. FacTweet performs better than the proposed baselines and obtains the highest macro-F1 value of $0.565$. Our results indicate the importance of taking into account the sequence of the tweets in the accounts' timelines. The sequence of these tweets is better captured by our proposed model sequence-agnostic or non-neural classifiers. Moreover, the results demonstrate that the features at tweet-level do not perform well to detect the Twitter accounts factuality, since they obtain a result near to the majority class ($0.18$). Another finding from our experiments shows that the performance of the Tweet2vec is weak. This demonstrates that tweets' hashtags are not informative to detect non-factual accounts. In Table \ref{table:ablation}, we present ablation tests so as to quantify the contribution of subset of features. The results indicate that most performance gains come from words embeddings, style, and morality features. Other features (emotion and sentiment) show lower importance: nevertheless, they still improve the overall system performance (on average 0.35\% Macro-F$_1$ improvement). These performance figures suggest that non-factual accounts use semantic and stylistic hidden signatures mostly while tweeting news, so as to be able to mislead the readers and behave as reputable (i.e., factual) sources. We leave a more fine-grained, diachronic analysis of semantic and stylistic features -- how semantic and stylistic signature evolve across time and change across the accounts' timelines -- for future work.


\begin{table}[t]
\caption{FacTweet: experimental results.}
\begin{subtable}[t]{.55\linewidth}
\caption{Results on account classification.\label{table:results}}
\begin{tabular}{l||cc}
  \hline
   \textbf{Methods} & Acc. & Macro-F1 \\
  \hline
  \multicolumn{3}{c}{\textbf{Baselines}} \\
  \hline
  Majority Class & { 0.56 } & { 0.18 }  \\
  \hline
  Random class & { 0.25 } & { 0.21 }  \\
  \hline
  Bag-of-Words & { 0.60 } & { 0.28 }  \\
  \hline
  Tweet2vec & { 0.56 } & { 0.18 }  \\
  \hline
  \multicolumn{3}{c}{\textbf{Tweet-level approaches}} \\
  \hline
  LR + All & { 0.67 } & { 0.39 } \\
  \hline
  LR + All (top-500 replied)  & { 0.69 } & { 0.41 } \\
  \hline
  LR + FacTweet & { 0.65 } & { 0.351 } \\
  \hline
  \multicolumn{3}{c}{\textbf{Chunk-level approaches}} \\
  \hline
  LR + All & { 0.737 } & { 0.559 }  \\
  \hline
  \hline
  \textbf{FacTweet} & { \textbf{0.74} } & { \textbf{ 0.565 } }  \\
\end{tabular}
\end{subtable}
\begin{subtable}[t]{.45\linewidth}
\caption{Ablation tests.\label{table:ablation}}
\begin{tabular}{p{3cm}||cc}
  \hline
  \textbf{Methods} & Acc. & Macro-F1 \\
  \hline
  LR + All    & { 0.737 } & { 0.559 }  \\
  \hline
  $-$ Emotion         & { 0.731 } & { 0.557 }  \\
  \hline
  $-$ Sentiment        & { 0.731 } & { 0.554 }  \\
  \hline
  $-$ Morality         & { 0.725 } & { 0.548 }  \\
  \hline
  $-$ Style            & { 0.737 } & { 0.514 }  \\
  \hline
  $-$ Words embeddings & { 0.678 } & { 0.437 }  \\
\end{tabular}
\end{subtable}
\end{table}


\section{Conclusions}
In this paper, we proposed a model that utilizes chunked timelines of tweets and a recurrent neural model in order to infer the factuality of a Twitter news account. Our experimental results indicate the importance of analyzing tweet stream into chunks, as well as the benefits of heterogeneous knowledge source (i.e., lexica as well as text) in order to capture factuality. In future work, we would like to extend this line of research with further in-depth analysis to understand the flow change of the used features in the accounts' streams. Moreover, we would like to take our approach one step further incorporating explicit temporal information, e.g., using timestamps. Crucially, we are also interested in developing a multilingual version of our approach, for instance by leveraging the now ubiquitous cross-lingual embeddings \cite{ghanem19,glavas19}.


%
%
\bibliographystyle{splncs04}
\bibliography{references}
\end{document}